\documentclass[letterpaper, 10pt, conference]{ieeeconf}  

\usepackage{times}  
\usepackage{helvet}  
\usepackage{courier}  
\usepackage[hyphens]{url}  
\usepackage{graphicx} 
\usepackage{microtype}
\usepackage{graphics}
\usepackage{booktabs} 
\usepackage{caption}

\usepackage{hyperref}
\usepackage{cite}

\usepackage{makecell}

\usepackage{epstopdf}
\usepackage{rotating,multirow}
\usepackage{amsmath,amssymb} 
\usepackage{color}
\usepackage{wrapfig,lipsum}
\usepackage{tabularx}

\usepackage{algorithm}
\usepackage{algorithmic}
\usepackage[algo2e]{algorithm2e}
\usepackage{mathtools}

\newcommand{\etal}{\textit{et al. }}
\newcommand{\eg}{\textit{e.g., }}
\newcommand{\ie}{\textit{i.e., }}
\newcommand{\aka}{\textit{a.k.a. }}

\makeatletter
\newcommand*\bigcdot{\mathpalette\bigcdot@{.5}}
\newcommand*\bigcdot@[2]{\mathbin{\vcenter{\hbox{\scalebox{#2}{$\m@th#1\bullet$}}}}}
\makeatother

\usepackage{algorithm}
\usepackage{algorithmic}
\usepackage{amsmath}
\usepackage{newfloat}
\usepackage{listings}

\IEEEoverridecommandlockouts                              

\title{\LARGE \bf Camera-Tracklet-Aware Contrastive Learning for\\Unsupervised Vehicle Re-Identification}

\author{Jongmin Yu$^{1}$, Junsik Kim$^{2}$, Minkyung Kim$^{2}$, and Hyeontaek Oh$^{1}$
\thanks{$^{1}$Institute for IT Convergence, Korea Advanced Institute of Science and Technology (KAIST), Daejeon, 34141, Republic of Korea,{\tt \{andrew.yu, hyeontaek\}@kaist.ac.kr}}
\thanks{$^{2}$ School of Electrical Engineering, Daejeon, KAIST, 34141, Republic of Korea,
{\tt mibastro@gmail.com, mkkim1778@kaist.ac.kr}}
\thanks{Email for corresponding author: hyeontaek@kaist.ac.kr}

}

\begin{document}

\maketitle
\thispagestyle{empty}
\pagestyle{empty}

\begin{abstract}
Recently, vehicle re-identification methods based on deep learning constitute remarkable achievement. However, this achievement requires large-scale and well-annotated datasets. In constructing the dataset, assigning globally available identities (Ids) to vehicles captured from a great number of cameras is labour-intensive, because it needs to consider their subtle appearance differences or viewpoint variations. In this paper, we propose camera-tracklet-aware contrastive learning (CTACL) using the multi-camera tracklet information without vehicle identity labels. The proposed CTACL divides an unlabelled domain, \ie entire vehicle images, into multiple camera-level subdomains and conducts contrastive learning within and beyond the subdomains. The positive and negative samples for contrastive learning are defined using tracklet Ids of each camera. Additionally, the domain adaptation across camera networks is introduced to improve the generalisation performance of learnt representations and alleviate the performance degradation resulted from the domain gap between the subdomains. We demonstrate the effectiveness of our approach on video-based and image-based vehicle Re-ID datasets. Experimental results show that the proposed method outperforms the recent state-of-the-art unsupervised vehicle Re-ID methods. The source code for this paper is publicly available on \url{https://github.com/andreYoo/CTAM-CTACL-VVReID.git}.
\end{abstract}

\section{Introduction}
\label{sec:intro}
Vehicle re-identification (re-id) is a task to identify the vehicles of the same identities across the cameras. It is an essential procedure in modern intelligent traffic management systems. The primary challenge for vehicle re-id is to derive a robust representation model that can cover various details of captured vehicle images and distinguish differences among them. In the past few years, with the rising of deep learning, supervised methods based on deep learning improve vehicle re-id performance dramatically \cite{liu2016deep,liu2016deep_db,wei2018person,wang2017orientation,sensetime17,spatio_temporal_path,liu2016deep}. Those methods achieved outstanding performance compared with previous hand-crafted feature-based methods \cite{liao2015person,zheng2015scalable}. However, supervised methods require a well-labelled dataset which may not be available on a large-scale.

To overcome the dependence of labelled datasets, various unsupervised learning approaches have been proposed \cite{huang2020dual,he2020multi,peng2020unsupervised,yu2021unsupervised}. The predominant approach is to use transfer learning or domain adaptation (DA), and those suggest training a model with a pre-labelled dataset (\aka source domain) before adapting to unlabelled datasets (\aka target domain) \cite{huang2020dual,he2020multi,peng2020unsupervised}. However, those approaches still require laborious annotations for the source domain and may fail when the domain gap is large \cite{wei2018person}.

In recent, fully unsupervised re-id methods \cite{fan2018unsupervised,Fu_2019_ICCV,wang2020unsupervised,lin2020unsupervised}, which employ a pseudo-label generation for creating supervisory signals to train their model, have been proposed. However, those approaches methodologically cannot perfectly filter false-positive prediction results, which can significantly degrade re-id performance, during the pseudo-label generation. However, those approaches methodologically cannot perfectly filter false-positive predictions during the pseudo-label generation, which can significantly degrade re-id performance.

Therefore, in this paper, we present camera-tracklet-aware contrastive learning (CTACL) to boost the performance of vehicle re-id without explicit vehicle identity labels or a pre-labelled source dataset, which only uses cameras and tracklet identities for reducing the risk of false-positive predictions results. Camera Ids is one of the general information contained in re-id datasets, and tracklet is cost-free by-product information since tracking is a commonly used function in collecting person or vehicle images on re-id studies \cite{wei2018person,veriwild,zheng2016mars,zhao2021phd}. Thus, this problem setting is fairly reasonable.

When unlabelled vehicle images, camera Ids, and tracklet Ids are given, we divide the entire images into camera-level subdomains and conduct contrastive learning to each subdomain. Positive and negative samples for contrastive learning are decided by using the tracklet Ids. Camera-tracklet-aware memory (CTAM) is introduced to store and manage extracted features and the Ids. Additionally, as the proposed contrastive learning mainly operates in each subdomain, we introduce a DA across cameras to prevent the model from learning a camera-specific representation.

Our model is mainly evaluated on the VVeRI-901 dataset \cite{zhao2021phd} which is the first video-based vehicle re-id evaluation benchmark. Additionally, we reorganise image vehicle re-ID datasets, \ie VeRi-776 dataset~\cite{liu2016deep} and VeRi-Wild dataset~\cite{wei2018person}, to simulate the video vehicle re-id scenario for the evaluation. Compared with existing state-of-the-art unsupervised vehicle re-id methods, including DA-based methods \cite{huang2020dual,he2020multi,peng2020unsupervised}, our method produces state-of-the-art vehicle re-id performance with large performance margins. Our method produces rank-1 accuracies of 89.3 and 38.2 on VeRi-776 and VVeRi-901 datasets, respectively. These are 11.9\% and 9.6\% improved achievements against the best performing competitors. The second-ranked methods are VACP-DA \cite{zheng2020aware} on the VeRi-776 with rank-1 accuracy of 77.4, and SSL \cite{lin2020unsupervised} on the  VVeRI-901 with rank-1 accuracy of 28.6. Consequently, the proposed method demonstrates that it can provide promising performance without any types of labelled data.

\begin{figure*}[t]
	\centering
	\includegraphics[width=\textwidth]{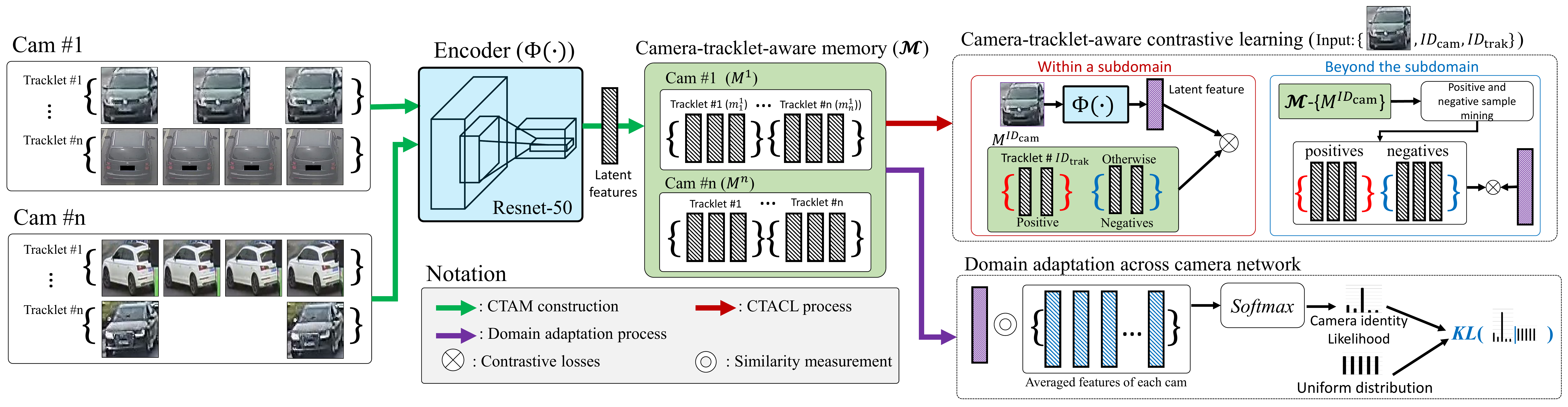}
	\caption{The training process of the proposed CTACL and DA using the CTAM. The positive and negative samples on the CTACL for each subdomain are definitively defined using tracklet Id ${ID}_{\text{Trak}}$ within the subdomain specified by camera Id ${ID}_{\text{Cam}}$. In this process, to improve the generalisation performance of the model, potential positive samples located beyond the subdomain are applied through positive sample mining. In addition, the DA across cameras is performed to improve the generalisation performance of the model explicitly.}
	\label{fig:2}
	\vspace{-2ex}
\end{figure*}

\section{Preliminary}
\subsection{Contrastive learning}
The contrastive learning is presented to learn such an embedding space in which similar sample pairs stay close to each other while dissimilar ones are far apart. Triplet loss \cite{schroff2015facenet}, noise contrast estimation (NCE) \cite{gutmann2010noise}, and InfoNCE \cite{oord2018representation} are well-known approaches among the contrastive learning. Recently, contrastive learning combined with self-supervised tasks is shown to be powerful in learning  a robust representation model with impressive performance in various visual recognition tasks \cite{chen2020simple,tian2020contrastive,khosla2020supervised}.

With an encoder ($\Phi(\cdot)$) and a single batch of images $\boldsymbol{x}=\{x_{i}\}_{i=1:B}$, where $B$ is the number of images in the batch, the contrastive learning is performed as a self-supervision manner (e.g., \cite{chen2020simple,tian2020contrastive,henaff2020data}), the loss is defined as follows:
\begin{equation}
  \mathcal{L}^{cl}=\sum_{i\in B}\mathcal{L}_i^{cl}
  =-\sum_{i\in B}\log{\frac{\text{exp}\left(z_i\bigcdot{}z_{i}^{+}/\tau \right)}{\sum\limits_{a\in A(i)}\text{exp}\left(\boldsymbol{z}_i\bigcdot\boldsymbol{z}_a/\tau\right)}
  }
  \label{eq:sscl}
\end{equation}
Here, $z=\Phi{}(x)\in\mathcal{R}^{D_\text{L}}$, where $D_\text{L}$ denotes the dimensionality of a latent feature space. The $\bigcdot$ symbol indicates the inner product. $\tau\in[0,1]$ is a scalar temperature parameter, and $A(i)\equiv B-\{i\}$. The $z_{i}$ is called the \emph{anchor} feature, $z_{i}^{+}$ is called the \emph{positive} features related to the anchor, and the others are all called the \emph{negatives} features. Positive features corresponded to the anchors are usually obtained by images augmented from anchor images \cite{khosla2020supervised}. Simple image augmentation techniques such as random cropping \cite{tian2020contrastive} and rotating \cite{henaff2020data} have been used.

\section{The proposed approach}
\label{sec:pm}
\subsection{Camera-tracklet-aware contrastive learning}
Our motivation is as follows: Even if identity labels for vehicle images, which can verify vehicle Ids between non-overlapping cameras, are not provided, if the camera and tracklet Ids are available, we can define multiple labelled subdomains by dividing unlabelled domain (\ie the entire image) into camera-level subdomains and by assigning temporal labels using tracklet. Based on this insight, we derive camera-tracklet-aware contrastive learning. Fig. \ref{fig:2} illustrates the entire workflow of the proposed approach.

We assume that a re-id model can use not only vehicle images $X=\{x_{i}\}_{i=1:n}$ but the Ids of cameras $Y_{\text{Cam}}=\{y^{\text{C}}_{i}\}_{i=1:n}$ and tracklet $Y_{\text{Trk}}=\{y^{\text{T}}_{i}\}_{i=1:n}$, where $n$ is the number of vehicle images. The proposed approach initially extracts latent feature $Z=\{z_{i}\}_{i=1:n}$ from each image using an encoder function $\Phi{}(\cdot):X\xrightarrow{}Z$. The extracted features are regularised by $l2$-normalisation to improve scale consistency of features in applying contrastive learning and DA.  

After extracting the features, a camera-tracklet-aware memory (CTAM) $\mathcal{M}$ is constructed. As shown in Fig. \ref{fig:2}, the CTAM is a memory bank storing the extracted features, and it is accessible by the camera and tracklet Ids. The CTAM is defined as follows:
\begin{equation}
\begin{aligned}
&\mathcal{M}=\{M^{i}\}^{i={1:\text{n}_{\text{Cam}}}}{} ,\\
&M^{i}=\{m^{i}_{j}\}_{j={1:\text{n}^{i}_{\text{Trk}}}}{} ,\\
&m^{i}_{j} = \{\hat{z}^{i,j}_{k}\}_{k=1:n^{i,j}_{\text{Img}}}{}.
  \label{eq:ctam}
 \end{aligned}
\end{equation}
where $M^{i}$ and $m^{i}_{j}$ denotes $i^{\text{th}}$ camera's feature sets and the $j^{\text{th}}$ tracklet's feature sets included in $M^{i}$, respectively.  $\text{n}_{\text{Cam}}$, $\text{n}^{i}_{\text{Trk}}$, and $n^{i,j}_{\text{Img}}$ denote the numbers of camera, tracklets on $i^{\text{th}}$ camera, and vehicle images in $j^{\text{th}}$ tracklet of $i^{\text{th}}$ camera, respectively. $\hat{z}$ is stored features in the CTAM.

During training, $\hat{z}$ is updated every training step as follow:
\begin{equation}
\begin{aligned}
\hat{z}^{t} = \frac{\hat{z}^{t-1}+z^{t}}{||\hat{z}^{t-1}+z^{t}||_{2}},  \quad \hat{z}^{0} = z^{0},
  \label{eq:update}
 \end{aligned}
\end{equation}
where $t$ indicates $t^{\text{th}}$ training step.

Based on the CTAM, we apply contrastive learning using camera and tracklet Ids (CTACL). One possible option is to directly apply self-supervised contrastive learning (SSCL) (Eq. \eqref{eq:sscl}) for vehicle re-id. However, it may be unsuitable for video re-id setting, since the loss function does not regard the existence of multiple positive samples. If a re-id model incorrectly considers actual positive samples to negative samples for training, the re-id performance would be significantly degraded.

As we aforementioned, by using camera and tracklet Ids, we can transform the unlabelled domain to a set of labelled subdomains.
By this transformation, although it is limited to each subdomain, we can deterministically distinguish whether a sample is positive or negative. Therefore, we propose a contrastive learning loss with multiple positive samples.

Given an unlablled vehicle image $x_{i}$ with camera $y_{i}^{\text{C}}$ and tracklet $y_{i}^{\text{T}}$ Ids, the loss function of the CTACL based on CTAM is defined as follows:
\begin{equation}
\begin{aligned}
  \mathcal{L}_{\text{CTACL}}=-\frac{1}{|m^{y^{\text{C}}_{i}}_{y^{\text{T}}_{i}}|}\sum_{\hat{z}_{p}\in m^{y^{\text{C}}_{i}}_{y^{\text{T}}_{i}}}\log{\frac{\text{exp}(z_i\bigcdot\hat{z}_p/\tau)}{\sum\limits_{\hat{z}_{a}\in M^{y^{\text{C}}_{i}}}\text{exp}(z_i\bigcdot\hat{z}_a/\tau)}},
  \label{eq:sub_ctacl}
\end{aligned}
\end{equation}
where $z_{i}$ is a latent feature extracted by $\Phi{}(x_{i})$. $\tau$ is the temperature parameter for contrastive learning, and $m^{y^{\text{C}}_{i}}_{y^{\text{T}}_{i}}$ is a set of features specified by the tracklet Id $y^{\text{T}}_{i}$ and the camera Id $y^{\text{C}}_{i}$. $|m^{y^{\text{C}}_{i}}_{y^{\text{T}}_{i}}|$ indicates the cardinality of the set $m^{y^{\text{C}}_{i}}_{y^{\text{T}}_{i}}$.

In computing the loss, the CTACL determines the subdomain for computing the loss through the camera Id: $M^{y^{\text{C}}_{i}}$, and it determines a set of positive samples using the tracklet Id: $m^{y^{\text{C}}_{i}}_{y^{\text{T}}_{i}}$. This means that only samples within the same subdomain are used to calculate the loss function. However, to derive robust vehicle re-id, it is essential to learn a generalised representation that can cover all other subdomains.

One straightforward way to overcome this issue is to interconnect subdomains by sampling the positive and negative samples from different subdomains for cross-domain contrastive learning. We define two types of positive samples: easy and hard positives. Among the samples of all other subdomains, the easy positive samples are defined by $k$-nearest samples of input images, as follows:
\begin{equation}
\begin{aligned}
&\text{P}^{\text{PS}}_{i} = \mathop{\arg \operatorname {sort}}_{\hat{z}_{j}\in\mathcal{M}-\{m^{y^{\text{C}}_{i}}_{y^{\text{T}}_{i}}\}} z_{i}\bigcdot{}\hat{z}_{j},\\
&\text{P}^{\text{Easy}}_{i} \longleftarrow{} \text{P}^{\text{PS}}_{i}[1:k],
\label{eq:easy} 
\end{aligned}
\end{equation}
where $\text{P}^{\text{PS}}_{i}$ is the sorted set of features based on pair-wise similarities between $z_{i}$ and other features stored in the CTAM ($\hat{z}_{j}$), and $\text{P}^{\text{Easy}}_{i}$ is the set of selected features as the easy positive. The easy positive samples mean samples taken from different cameras, which are similar to the input sample.

On the other hand, the hard positive samples mean the samples taken from different cameras, which are maximally different from the input sample. The hard positive samples are defined by the $k$-nearest samples of the samples having the same tracklet Id but the farthest from the input sample in the feature space, represented as follows:
\begin{equation}
\begin{aligned}
&\bar{z}' = \mathop{\arg \operatorname{min}}_{\bar{z}\in{}m^{y^{\text{C}}_{i}}_{y^{\text{T}}_{i}}} z_{i}\bigcdot{}\hat{z}_{j},\\
&\text{P}^{\text{PS}}_{\hat{z}'} = \mathop{\arg \operatorname {sort}}_{\hat{z}_{j}\in\mathcal{M}-\{m^{y^{\text{C}}_{i}}_{y^{\text{T}}_{i}}\}} \bar{z}'\bigcdot{}\hat{z}_{j},\\
&\text{P}^{\text{Hard}}_{i} \longleftarrow{} \text{P}^{\text{PS}}_{\hat{z}'}[1:k],
\label{eq:hard} 
\end{aligned}
\end{equation}
where $\bar{z}'$ indicates the sample having the same tracklet Id and the farthest from the input sample. $\text{P}^{\text{PS}}_{\hat{z}'}$ and $\text{P}^{\text{Hard}}_{i}$ are the sorted set of features using $\bar{z}'$ and the set of selected features as the hard positives, respectively. 

In selecting negative samples from other subdomains, we establish a grey zone that defines a non-obtainable area in acquiring negative samples to reduce the possibility of false-negative selection. After positive sample mining is completed, the remaining samples are sorted in descending order using a similarity score. And then, among the sorted features, samples corresponding to $\gamma$\% of the remaining samples are excluded from the top, and the remainders are designated as negative samples: $\text{N}_{i} \leftarrow{} \text{P}^{\text{PS}}_{i} [\lceil\gamma|\text{P}^{\text{PS}}_{i}|\rceil:|\text{P}^{\text{PS}}_{i}|]$, where $\lceil\cdot{}\rceil$ denotes ceiling function. 

By using the above two positive sample sets ($\text{P}^{\text{Easy}}$ and $\text{P}^{\text{Hard}}$) and the negative samples ($\text{N}_{i}$), we extend the CTACL as follows:
\begin{equation}
\begin{aligned}
  &\mathcal{L}_{\text{CTACL}}^{\text{Extend}}=\\
  &-\frac{1}{|m^{y^{\text{C}}_{i}}_{y^{\text{T}}_{i}}|+|\text{P}^{\text{+}}_{i}|}\sum_{\hat{z}_{p}\in m^{y^{\text{C}}_{i}}_{y^{\text{T}}_{i}}+\text{P}^{+}_{i}}\log{\frac{\text{exp}(z_{i}\bigcdot\hat{z}_p/\tau)}{\sum\limits_{\hat{z}_{a}\in M^{i}+\text{P}^{+}_{i}+\text{N}_{i}}\text{exp}(z_i\bigcdot\hat{z}_a/\tau)}},
  \label{eq:ext_ctacl}
\end{aligned}
\end{equation}
where $\text{P}^{\text{+}}_{i}$ denotes the union set of the two positive sample sets $\text{P}^{\text{Easy}}$ and $\text{P}^{\text{Hard}}$. $\text{N}_{i}$ represents the negative sample set.

CTACL using CTAM enables generalised non-parametric representation learning on the variational domain problem. For example, the number of tracklets and the images on each tracklet can vary. When a parametric approach (\eg classification based on neural networks) is used, the parameter setting such as the output dimensions should be revised. However, CTACL using CTAM can provide a learning approach without any parametric model, which gives better methodological flexibility.

\subsection{Domain adaptation across camera networks}
The CTACL includes a process for taking positive and negative samples from out of subdomains to improve generalisation performance. However, it may not provide sufficient generalisation performance for the entire camera network. It is a strong likelihood that the number of selected positive samples by the mining would be much smaller compared with the true positive samples. We can manually increase $k$ to take positive samples from beyond each subdomain as many as possible, but this approach also can generate a great number of false positives, which can degrade the re-id model performance.

To resolve this issue, we propose camera-level DA using CTAM to explicitly improve the generalisation performance of the learnt representation. As shown in Fig. \ref{fig:2}, the proposed DA aims to uniformise the likelihood of camera Id classification. Intuitively, uniformising the likelihood of camera Id classification can be interpreted as maximising uncertainty of camera-specified information, and it can be considered as reducing the informational bias obtained in each subdomain.

To do this, we define the likelihood that the input image $x$ will be classified by the $i^{\text{th}}$ camera Id as follows:
\begin{equation}
\begin{aligned}
P(y^{\text{C}}=i|z) = \frac{\text{exp}(z\bigcdot{}\bar{z}^{i})}{\sum_{j=1}^{n_{\text{Cam}}}\text{exp}(z\bigcdot{}\bar{z}^{j})},
  \label{eq:cls_cids}
\end{aligned}
\end{equation}
where $z$ is the latent feature obtained by $\Phi{}(x)$, and $\bar{z}^{i}$ indicates the centre point of feature distribution having $i^{\text{th}}$ camera Id, and it is defined by the average of the features having the same camera Id as follow: \begin{equation}
\begin{aligned}
\bar{z}^{i} = \frac{1}{|M^{i}|}\sum_{j=1}^{n^{i}_{\text{Trk}}}\sum_{k=1}^{n^{i,j}_{\text{Img}}}\hat{z}^{i,j}_{k},
\end{aligned}
\end{equation}
where $|M^{i}|$ denotes the number of features classified by $i^{\text{th}}$ camera Id. As same as all other features, the averaged features are also $l2$-normalised after computing it.

The loss function for the DA between cameras is formulated based on Kullback–Leibler (KL) divergence between the likelihoods and the uniform distribution, which is defined as follows:
\begin{equation}
\begin{aligned}
\mathcal{L}_{\text{DA}} =KL(U(y^{\text{C}})||P(y^{\text{C}}|z)) = \sum_{i}^{n_{\text{Cam}}}U(y^{\text{C}}={i})\frac{U(y^{\text{C}}=i)}{P(y^{\text{C}}=i|z)}.
  \label{eq:da_kld}
\end{aligned}
\end{equation}
Here, the uniform distribution vector $U(y^{\text{C}})$ is defined according to the number of cameras as follows:
\begin{equation}
\begin{aligned}
U(y^{\text{C}}=i)=\frac{1}{n_{\text{Cam}}}.
  \label{eq:uniform}
\end{aligned}
\end{equation}
By minimising the KL-divergence between the uniform distribution and the likelihoods for camera Id classification, our re-id method maximise uncertainty in distinguishing specific camera Id, and subsequently, it reduces the risk of subdomain-specific representation learning and improve generalisation performance. 

The objective function for joint learning using the CTACL and the proposed DA is defined as follows:
\begin{equation}
\begin{aligned}
\mathcal{L}_{\text{Total}} = \mathcal{L}_{\text{CTACL}}^{\text{Extend}}+\lambda \mathcal{L}_{\text{DA}},
  \label{eq:total}
\end{aligned}
\end{equation}
where $\lambda$ indicates a balancing weight for the DA loss.

\begin{table}
\resizebox{\columnwidth}{!}{%
\begin{tabular}{l|c|c|c|c|c}
\hline
Dataset & Imgs &  Ids & Imgs$/$Id &  Cams & Time (H)\\
\hline\hline
VVeRI-901 &  488,195 & 901 & 541.83  & 11 & 33\\
\hline
Veri-Wild &  416,314  &  40,671  &10.23 & 174 & 125,280\\
\hline
VeRi-776 & 49,360 & 776 & 63.60  & 18 & 18\\
\hline
\end{tabular}
}
\caption{Key properties of the vehicle re-id datasets. `Imgs', `Ids', `Cams', and `Time' denote the number of images, vehicle identities, cameras, and recording time of each dataset, respectively.}
\vspace{-2ex}
\label{tbl:dataset_explain}
\end{table}

\section{Experiments}
\subsection{Dataset and Evaluation metrics}
We use the three publicly available datasets: VVeRI-901 \cite{zhao2021phd}, VeRi-776 \cite{liu2016deep}, and Veri-Wild \cite{wei2018person} on the ablation study and performance comparison with recent state-of-the-art methods for unsupervised vehicle re-id. Table \ref{tbl:dataset_explain} shows the key properties of those datasets. Since we assume that the proposed approach can leverage camera and tracklet Ids, the VVeRI-901 dataset, which is proposed to provide a benchmark for video-based vehicle re-id task, is used as the main benchmark to evaluate our methods.

Additionally, we reorganised VeRi-776 and Veri-Wild datasets, which are image-based vehicle re-id datasets, to assign virtual tracklet Ids to each sample. Since those two datasets already provide camera Ids, we additionally create the virtual tracklet Ids by mapping the vehicle IDs included in each camera to a new Ids that are not shared between cameras in order to secure uniqueness of the tracklet Ids across the entire camera network. Our experiments are conducted based on the standard experimental protocol for unsupervised vehicle re-id  \cite{liu2016deep,liu2016deep_db,wei2018person,yu2021unsupervised}. Cumulative Matching Characteristics (CMC) and Mean Average Precision (mAP) are used as performance evaluation metrics.

\begin{figure}[t]
    \centering
   \includegraphics[width=\columnwidth]{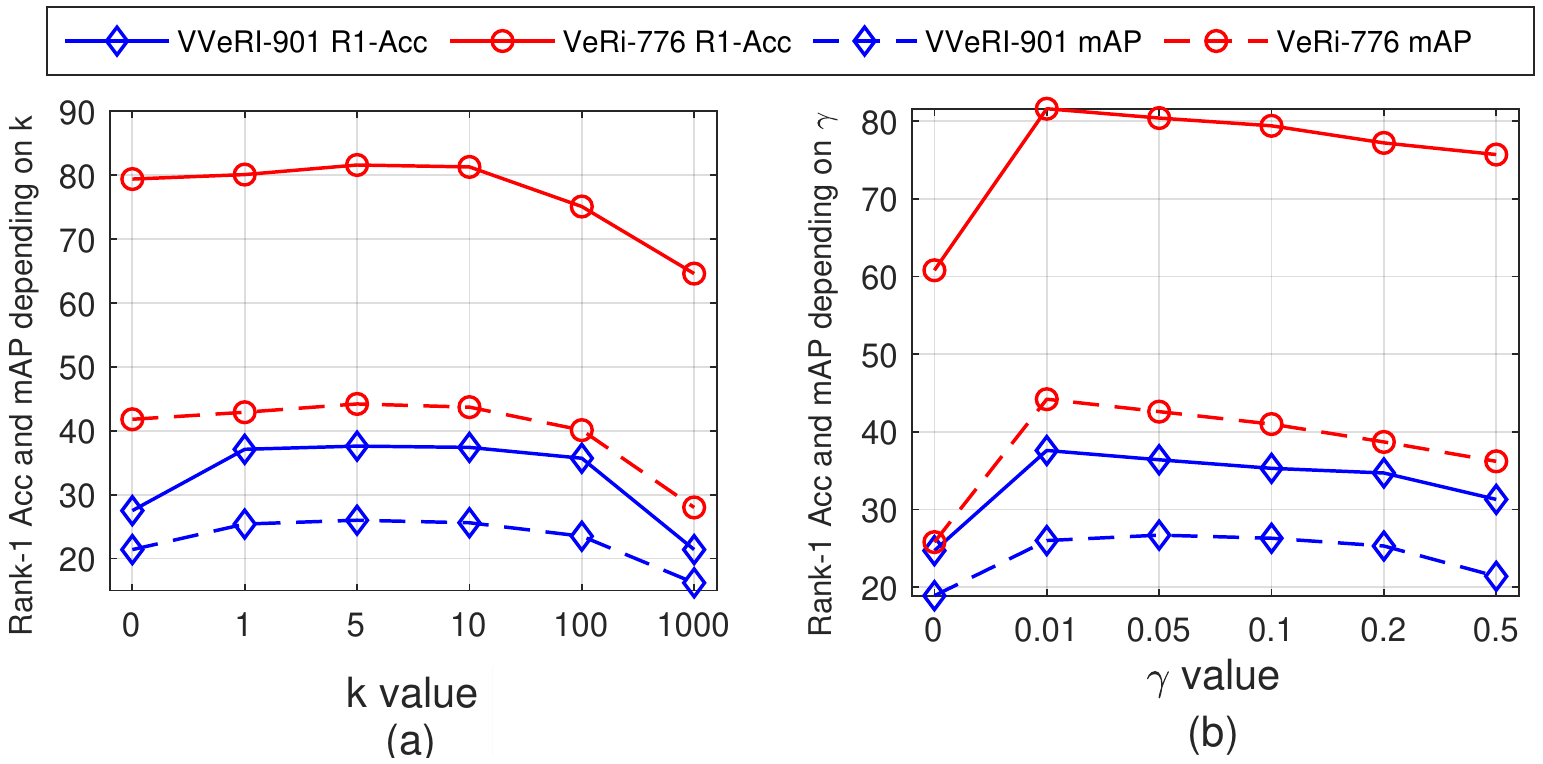}
    \caption{Performance analysis depending on the values of $k$ and $\gamma$ on VVeRI-901 and VeRi-776 datasets. (a) shows the rank-1 accuracies and mAP depending on $k$. (b) represents the trends of rank-1 accuracies and mAP according to $\gamma$.}
\label{fig:kg}  
\vspace{-2ex}
\end{figure}

\subsection{Implementation}
 \label{sec:exp:2}
All images are resized to 256$\times$128. Stochastic gradient descent (SGD) with a momentum of 0.9 is used for model optimisation. The training epoch is set by 50. At the beginning of model training, the learning rate is set by 0.1 and decayed by multiplying 0.1 for every 10 epoch. The batch size is 256. ResNet-50 \cite{he2016deep} followed by $l2$-normalisation layer is used as the encoder function. The encoder model is pre-trained by ImageNet \cite{krizhevsky2012imagenet}. The output dimensionality of the encoder function is 2,048. We used simple data augmentations (such as random crop, rotation, and colour jitters) to boost the generality of learnt representations. Initially, the CTACL is applied within subdomains (Eq. \eqref{eq:sub_ctacl}) for 5 epochs to ensure the minimum representation learning performance; after that the extended CTACL (Eq. \eqref{eq:ext_ctacl}) with the DA is applied. CTAM is completely overhauled every 5-epoch in training to improve the consistency between features. By referring to the research results of Wang \etal \cite{wang2021understanding} and Khosla \etal \cite{khosla2020supervised}, the temperature parameter $\tau$ is set to 0.07. The balancing weight $\lambda$ and the grey zone scale $\gamma$ are fixed by 0.2 and 0.01, respectively (based on the ablation study).

\subsection{Ablation Study}
\label{sec:exp:3}
The performance of our method is affected by the following three hyper-parameters: $k$, $\gamma$, and $\lambda$. We observe performance changes according to the variations of those parameters. We also demonstrate the effectiveness of the CTACL and DA by comparing them with various loss functions. All experiments are conducted with unsupervised vehicle re-id settings on the VVeRI-901 and VeRi-776 datasets. Parameters that are not subject to monitoring are fixed during the experiments. The parameter achieving the best performance would be fixed for further experiments.

\textbf{Parameter analysis on $k$}: The $k$ in Eq. \eqref{eq:easy} and Eq. \eqref{eq:hard} decides how many potential positive samples to be selected from outside of subdomains. We observe the rank-1 accuracy and mAP in terms of $k$. The experimental results in Fig. \ref{fig:kg}(a) show, in the interval where $k$ value increases from 0 to 1000, the performance increases rapidly, and after that, the performance is gradually decreased. These experimental results show that using the positive samples mined from other subdomains can improve the vehicle re-id performances by giving a chance to learn more general representation. However, when $k$ is getting larger, the possibility of wrong mined results also being increased so, it degrades the performance. The best performance is achieved by $k$ of 5.

\textbf{Parameter analysis on $\gamma$}: $\gamma$ decides the range of the grey zone, which is the skipping area in selecting positive and negative samples from beyond subdomains. As shown in Fig \ref{fig:kg}(b), when the grey zone is not considered, \ie $\gamma$ is 0, the performances are lowest. The best performances are achieved by 0.01 of $\gamma$. The performance is slightly decreased when $\gamma$ is getting larger, but it is not significant compared with the performance increment between 0 to 0.01 of $\gamma$. The trends of the rank-1 accuracy and mAP shown in Fig. \ref{fig:kg}(b) can be interpreted that the advantage of contrastive learning using weak supervisory signals, which are camera and tracklet Ids, can be degraded by false-positive. The presence of grey areas has a significant impact on performance, but its size does not significantly affect performance.

\begin{table}[t]
\resizebox{\columnwidth}{!}{%
\begin{tabular}{l|c|c|c|c|c|c|c}
\hline
\multicolumn{2}{c|}{$\lambda$ setting}  & 0.0 & 0.01 & 0.1 & 0.2 & 0.5 & 1.0  \\
\hline
\multirow{2}{*}{VeRi-776} &Rank-1 & 81.0 & 87.4 & 89.1 & \textbf{89.3} & 88.7 & 88.2 \\
\cline{2-8} 
&mAP & 43.9  & 53.6 & 55.4 & \textbf{55.2}  & 54.8 & 54.9  \\
\hline
\multirow{2}{*}{VVeRI-901} &Rank-1 & 33.6 & 36.7 & 36.8 & \textbf{38.2} & 37.5 & 37.3 \\
\cline{2-8} 
&mAP & 25.8  & 26.6 & 26.7 & \textbf{29.0}  & 28.2 & 28.5  \\
\hline
\end{tabular}
}
\caption{Performance analysis depending on the setting of $\lambda$. The \textbf{bolded} figures denote the best performance.} 
\vspace{-2ex}
\label{tbl:comparison-lambda}
\end{table}

\begin{table}[t]
\resizebox{\columnwidth}{!}{%
\begin{tabular}{l|c|c|c|c}
\hline
\multirow{2}{*}{Loss function}  & \multicolumn{2}{c|}{VeRi-776} & \multicolumn{2}{c}{VVeRI-901}  \\
\cline{2-5} 
& Rank-1  & mAP  & Rank-1  & mAP    \\
\hline
SSCL (Eq. \ref{eq:sscl})  & 23.6 & 10.4 & 26.1 & 15.9 \\
\hline
Softmax CE+GT  & 94.8 & 79.8 & 43.5 & 42.8 \\
Softmax CE+Tracklets  & 50.8 & 16.1 & 28.6 & 16.5 \\
\hline
CTACL & 81.6 & 44.2 & 33.7 & 26.0 \\
CTACL+DA & 89.3 & 55.2 & 38.2 & 28.1\\
CTACL+GT & 92.3 & 68.2 & 41.1 & 29.5 \\
CTACL+DA+GT & 92.8 & 70.7 & 41.8 & 29.9 \\
\hline
\end{tabular}
}
\caption{Performance comparison of various loss function settings. Self-supervised contrastive learning (SSCL) and Softmax cross-entropy (CE) are used. `GT' indicates the model is trained by using the vehicle class labels. `Tracklet' denotes that the tracklets of each subdomain is used as a label. `DA' means the domain adaptation has been applied in the training phase.} 
\vspace{-2ex}
\label{tbl:comparison-loss}
\end{table}

\textbf{Effectiveness of DA with $\lambda$}: When $\lambda$ is 0, it means that the DA would not be considered during model training. On the other hand, when $\lambda$ is 1.0, it means that the gradient of the DA loss would be equally considered with the gradient of the contrastive learning loss (Eq. \eqref{eq:ext_ctacl}). We conduct the performance evaluation with six different values of $\lambda$ between 0 to 1, and the results is shown in Table \ref{tbl:comparison-lambda}. 

The best performance is obtained by 0.2 of $\lambda$, and it shows 89.3 of rank-1 accuracy and 55.2 of mAP on the VeRi-776 dataset and 38.2 of rank-1 accuracy and 29.0 of mAP on the VVeRI-901 dataset. The lowest performance on our experiments is obtained by 0 of $\lambda$, which means the DA is not applied. The experimental results on $\lambda$ show that the domain adaptation across the camera networks improves the vehicle re-id performance.

\begin{table}
\resizebox{\columnwidth}{!}{
\begin{tabular}{l|c|c|c|c|c}
\hline
\multirow{2}{*}{Method} & \multirow{2}{*}{Settings} & \multicolumn{4}{c}{VVeRI-901} \\
\cline{3-6}
& &Rank-1 & Rank-5 & Rank-10 & mAP  \\
\hline\hline
GoogLeNet~\cite{wei2018person} & SU &40.8 & 59.6& 65.3& 41.4 \\
ID Loss~\cite{zhong2019invariance} & SU & 37.2 &52.4& 60.5 & 36.5 \\
TCLNET-tri~\cite{Fu_2019_ICCV} & SU & 45.5 &58.0& 67.1 & 44.0 \\
MGH~\cite{yan2020learning} & SU & 44.3 & 61.8 & 67.8 & 44.5\\
Triplet~\cite{zhao2020unsupervised} & SU & 35.6 & 51.2 & 54.3 & 33.7 \\
PhD~\cite{zhao2021phd} & SU & \underline{47.1} & \underline{67.6} & \underline{74.7} & \underline{47.2}\\
\hline\hline
BOW~\cite{zheng2015scalable} & UN & 13.2 & 15.1 & 20.1 & 3.7\\
BUC~\cite{lin2019bottom} & UN & 15.2 & 17.6 & 23.7 & 5.8 \\
SSL~\cite{lin2020unsupervised} & UN & 28.6 & 36.8 & 38.1 & 19.2 \\
MMLP~\cite{wang2020unsupervised} & UN & 26.3 & 39.1 & 40.2 & 18.1 \\
SSML~\cite{yu2021unsupervised} & UN & 27.9 & 36.7 & 39.6 & 19.6 \\
\hline
CTACL  & UN & 33.7 & 37.6 & 40.3 & 26.0 \\
CTACL+DA  & UN & \textbf{38.2} & \textbf{42.1} & \textbf{43.4} & \textbf{29.0} \\
\hline
\end{tabular}
}
\caption{Comparison of person re-id performance on \textit{VVeRI-901}. `SU' and `UN' indicate a methods is based on supervised and fully unsupervised learning. The \underline{underlined} figures and \textbf{bolded} figures indicate the best performance among the supervised learning-based and unsupervised learning-based methods, respectively.} 
\vspace{-2ex}
\label{tbl:comparison-vveri}
\end{table}

\begin{table*}[t]
\begin{center}
\resizebox{\textwidth}{!}{  
\setlength{\tabcolsep}{3pt}
\begin{tabular}{l|c| c|c| c c c | c c c| c  c c|c c c}
\hline
\multirow{2}{*}{Methods} & \multirow{2}{*}{Year} & \multirow{2}{*}{Settings} & \multirow{2}{*}{Source}  & \multicolumn{3}{c|}{VeRi-776} &  \multicolumn{3}{c|}{Veri-Wild (Small)} & \multicolumn{3}{c|}{Veri-Wild (Medium)} & \multicolumn{3}{c}{Veri-Wild (Large)}\\
 \cline{5-7} \cline{7-16}
& & & & Rank-1  & Rank-5  & mAP & Rank-1& Rank-5 & mAP  & Rank-1 & Rank-5 & mAP & Rank-1  & Rank-5  & mAP \\
\hline\hline
SPGAN \cite{deng2018image} & 2018  & DA & VehicleID  & 57.4 & 70.0 & 16.4  &     59.1     &     76.2     &     24.1     &     55.0     &   74.5       &    21.6      &     47.4     &     66.1     &    17.5  \\
VR-PROUD \cite{bashir2019vr} & 2019 & DA  & VehicleID &  55.7 & 70.0 & 22.7 & -  & - & - & - & - & - & - & - & - \\
ECN \cite{zhong2019invariance} & 2019 & DA &  VehicleID & 60.8 & 70.9 & 27.7 &   73.4       &     88.8     &     34.7     &   68.6       &     84.6     &    30.6      &   61.0       &     78.2     &    24.7 \\
PAL \cite{he2019part}& 2020 & DA  & VehicleID&  68.2 &  79.9 & \underline{42.0} & -  & - & - & - & - & - & - & - & - \\
UDAP \cite{song2020unsupervised}& 2020 & DA  & VehicleID & 76.9 & \underline{85.8} & 35.8  &     68.4    &     85.3     &    30.0      &    62.5      &     81.8     &    26.2      &     53.7     &       73.9 &     20.8  \\
VACP-DA \cite{zheng2020aware}& 2020 & DA & VehicleID  &   \underline{77.4} & 84.6 & 40.3 & \underline{75.3} & \underline{89.0} & \underline{39.7} & \underline{69.0} & \underline{85.5} & \underline{34.5} & \underline{61.0} & \underline{79.7} & \underline{27.4} \\
AE \cite{journals/tomccap/DingFXY20} & 2020  & DA & VehicleID & 73.4 & 82.5 & 26.2  &     68.5     &     87.0     &  29.9        &     61.8     &     81.5     &   26.2       &     53.1     &    73.7      &    20.9  \\
\hline\hline
LOMO$^{\ddagger}$ \cite{liao2015person} & 2015  & UN &  - &  42.1 & 62.2 & 12.2   & 25.7  & 44.7  & 8.9 & 23.6 & 40.6  &   8.1 &  18.8 & 34.4& 5.9  \\ 
BOW$^{\ddagger}$ \cite{zheng2015scalable} & 2015  & UN & -     & 44.7    & 66.4  &  14.5   &    28.5     &   43.6       &     9.4     &     25.4     &     40.7     &    8.6      &     18.3    &    38.6      &   6.6 \\
BUC \cite{lin2019bottom}& 2019  & UN & -     & 54.7    & 70.4  &  21.2    &     37.5     &   53.0       &     15.2     &     33.8     &     51.1     &    14.8      &     25.2     &    41.6      &   9.2 \\
SSL \cite{lin2020unsupervised} & 2020  & UN & -     & 69.3    & 72.1  &  23.8  &     38.5     &   58.1       &     16.1     &     36.4     &     56.0     &    17.9      &     32.7     &    48.2      &   13.6 \\
MMLP \cite{wang2020unsupervised} & 2020  & UN & -     & 71.8    & 75.9  &  24.2  &     40.1     &   63.5       &     15.9     &     39.1     &     60.4     &    19.2      &     33.1     &    50.4      &   14.1 \\
SSML \cite{yu2021unsupervised} & 2021 & UN &  - &  \underline{74.5} & \underline{80.3} & \underline{26.7}  & \underline{49.6} & \underline{71.0} & \underline{23.7} & \underline{43.9}  & \underline{64.9} & \underline{20.4} & \underline{34.7} & \underline{55.4} & \underline{15.8}\\\hline \hline
CTACL & 2021 & UN &  - &  81.6 & 89.5 & 44.2  & 71.05 & 86.6 & 58.2 & 69.2  & 83.7 & 49.2 & 60.1 & 81.5 & 41.2\\
CTACL+DA & 2021 & UN &  - &  \textbf{89.3} & \textbf{93.9} & \textbf{55.2}  & \textbf{79.2} & \textbf{93.6} & \textbf{65.0} & \textbf{73.1}  & \textbf{89.5} & \textbf{56.2} & \textbf{63.6} & \textbf{83.5} & \textbf{44.9}\\
\hline
\end{tabular}
}
\end{center}
\caption{Performance comparison on unsupervised vehicle re-id with state-of-the-art methods on the VeRi-776 dataset~\cite{liu2016deep} and the Veri-Wild dataset~\cite{wei2018person}. `DA' and `UN' denote a method is based on domain-adaptation or fully unsupervised learning. `-' denotes that the results are not provided. The underlined results indicate the best performance among the DA-based methods. The \textbf{bolded} results indicate the best performance on the comparison.}
\label{tbl:comparison}
\vspace{-2ex}
\end{table*}

\textbf{Effectiveness of CTACL}: We compare the CTACL with softmax cross-entropy (CE) loss and the SSCL loss (Eq. \eqref{eq:sscl}). Based on the softmax CE, we derive the vehicle re-id models using the ground truth \ie vehicle class labels, and the tracklets. Also, we evaluate the performance of the CTACL trained by the ground truth. In this case, samples having the same vehicle class label are considered as the positive samples, and the remaining samples are all negatives; \ie the positive sample mining was not used. Table \ref{tbl:comparison-loss} contains the experimental results for this ablation study. 

The lowest performance is obtained by SSCL loss. The best performance is achieved by the softmax CE with the ground truth. It shows rank-1 accuracy of 94.8 and mAP of 79.8 on the VeRi-776 dataset, and it shows rank-1 accuracy of 43.5 and mAP of 42.8 on the VeRi-776 dataset. The CTACLs using the ground truth also shows similar performances. However, the experimental results show that when only a limited supervisory signal (\eg tracklet Ids) is available, softmax CE may not be a suitable solution. In contrast to the CTACL produces over than rank-1 accuracy of over 80 without explicit labels for vehicle Ids, the performances of the vehicle re-id model trained by softmax CE (rank-1 accuracy of 50.8) are significantly dropped when only tracklet Ids are given.

\subsection{Comparison with state-of-the-art methods}
\label{sec:exp:4}
We compare the CTACL with various state-of-the-art methods of unsupervised vehicle re-id. Unfortunately, only a few fully unsupervised vehicle re-id methods have been proposed and reported their performances. In particular, on the VVeRI-901 dataset, only the performances of supervised methods have been reported. Accordingly, we conducted additional experiments with several methods recently presented for fully unsupervised person re-id. Following studies are used for the performance comparison: LOMO  \cite{liao2015person}, BOW \cite{zheng2015scalable}, BUC \cite{lin2019bottom}, SSL \cite{lin2020unsupervised}, MMLP\cite{wang2020unsupervised}, and SSML \cite{yu2021unsupervised}. Those studies provide publicly available source codes, so the performance evaluation is carried out with those source codes.

\textbf{VVeRI-901 dataset}: Table \ref{tbl:comparison-vveri} shows the quantitative performance comparison using VVeRI-901 dataset. The CTACL with DA achieves rank-1 accuracy of 38.2 and mAP of 29.0. Those are the best performance among the unsupervised methods. The CTACL outperforms other unsupervised vehicle re-id methods with a minimum performance gap of 9.6\%. In comparison with supervised vehicle re-id methods, the performance of the CTACL is higher than several supervised learning-based models \cite{zhong2019invariance,zhao2020unsupervised}. The best performance on the VVeRI-901 dataset is achieved by PhD learning \cite{zhao2021phd}, and it shows about 8.9\% better performance compared to ours. 

The experimental results on the VVeRI-901 dataset can be interpreted as follows. Since the VVeRI-901 dataset contains much noisy information such as occlusion between vehicle detection results, the performances of the unsupervised vehicle re-id methods show inferior performance compared with supervised vehicle re-id methods. Obviously, using vehicle Ids is a clear advantage in deriving a robust vehicle re-id model. However, even though explicit vehicle Ids are not given, the CTACL can obtain promising performance by using contrastive learning with weak-supervisory signals such as camera and tracklet Ids. 

\textbf{VeRi-771 and Veri-Wild datasets}: The experimental results on VeRi-776 and Veri-Wild datasets also demonstrate the effectiveness of the CTACL. Table \ref{tbl:comparison} shows that the quantitative performance comparison between the CTACL and other unsupervised methods on Veri-776 and Veri-Wild datasets. The CTACL with DA achieves state-of-the-art performances on our experiments. The CTACL achieves rank-1 accuracy of 89.3 and mAP of 55.2 on the Veri-776 dataset. For the Veri-Wild dataset, it achieves rank-1 accuracy of 79.2 and mAP of 65.0. For the `Medium' and `Large' test sets, it produces rank-1 accuracy of 73.1 and mAP of 56.2 and rank-1 accuracy of 63.6 and mAP of 44.9, respectively. These figures are minimum of 2.9\% improved performance over the VACP-DA \cite{zheng2020aware} which is the DA-based method with rank 2.

In comparison with the fully unsupervised methods \cite{lin2019bottom,lin2020unsupervised,wang2020unsupervised,yu2021unsupervised}, the CTACL outperforms other fully unsupervised methods with significantly large margins. SSML \cite{yu2021unsupervised}, which is the second-ranked method among the fully unsupervised methods, produces rank-1 accuracy of 49.6 and mAP of 23.7 on the `Small' test set, rank-1 accuracy of 43.9 and mAP of 20.4 on the `Medium' test set, and rank-1 accuracy of 34.7 and mAP of 15.8 on the `Large' test set.

The overall experimental results demonstrate that the CTACL not only outperforms the existing state-of-the-art unsupervised vehicle re-id methods but also achieves comparable performance to the supervised learning-based methods. Additionally, performance comparisons between the CTACL and DA-based methods justify that if weak-supervisory signals are available, we can derive a robust vehicle re-id method without a labelled source dataset. 

\section{Conclusion}
\label{sec:con}
In this paper, we have proposed camera-tracklet-aware contrastive learning (CTACL). The proposed CTACL uses camera and tracklet information, which are easily obtained when a vehicle re-id dataset is constructed. Based on these two Ids, we divide an unlabelled domain (\ie the entire images), into multiple camera-level subdomains. Tracklet Ids corresponding to each subdomain are used to decide positive and negative samples to compute the CTACL loss. Also, we have applied the DA across camera networks to improve the generalisation performance of learnt representation. The ablation studies have demonstrated the effectiveness of CTACL and DA in boosting the unsupervised vehicle re-id performance. In comparison with various existing state-of-the-art methods on unsupervised vehicle re-id, the CTACL has outperformed other unsupervised methods, including the DA-based method.

\small
\bibliographystyle{IEEEtran}
\bibliography{icra_bib}

\end{document}